\pdfoutput=1
\documentclass[11pt]{article}

\usepackage[final]{acl}

\usepackage{times}
\usepackage{latexsym}

\usepackage[T1]{fontenc}
\usepackage[utf8]{inputenc}

\usepackage{microtype}

\usepackage{inconsolata}

\usepackage{graphicx}
\usepackage{xspace}
\usepackage{bm}
\usepackage{booktabs}
\usepackage{amsmath}
\usepackage{enumerate}
\usepackage{enumitem}
\usepackage{multirow}
\usepackage{threeparttable}
\usepackage{bbding}
\usepackage{algorithm}
\usepackage{algorithmic}

\def\fig{Figure.\xspace}

\def\sec{Sec.\xspace}
\def\tab{Table.\xspace}
\def\apx{Apx.\xspace}
\def\alg{Alg.\xspace}

\def\eg{{\textit{e.g.}\xspace}}

\def\etc{{\textit{etc}\xspace}}

\newcommand{\head}[1]{{\noindent \textbf{#1:}}}

\ifodd 0
\newcommand{\com}[1]{\textbf{\color{red}(COMMENT: #1)}} %comment of the text
\newcommand{\todo}[1]{\textbf{{\color{orange}(TODO: #1)}}}
\newcommand{\unused}[1]{{\color{gray}#1}}
\newcommand{\sheng}[1]{\textbf{\color{olive}(Sheng: #1)}} %comment of the text
\else

\newcommand{\com}[1]{}
\newcommand{\todo}[1]{}
\newcommand{\unused}[1]{}
\newcommand{\sheng}[1]{}

\fi

\title{ProMind-LLM: Proactive Mental Health Care via Causal Reasoning \\ with Sensor Data}

\author{
Xinzhe Zheng$^{3}$\thanks{Equal contribution.} \quad
Sijie Ji$^{1,2}$\footnotemark[1] \quad
Jiawei Sun$^{4}$\footnotemark[1] \quad
Renqi Chen$^{5}$ \quad \\
\textbf{Wei Gao}$^1$ \quad
\textbf{Mani Srivastava}$^2$ \\
$^1$California Institute of Technology \quad
$^2$UCLA \quad
$^3$National University of Singapore\\
$^4$Hangzhou Dianzi University \quad
$^5$Fudan University
}

\begin{document}
\maketitle
\begin{abstract}
Mental health risk is a critical global public health challenge, necessitating innovative and reliable assessment methods.
With the development of large language models (LLMs), they stand out to be a promising tool for explainable mental health care applications.
Nevertheless, existing approaches predominantly rely on subjective textual mental records, which can be distorted by inherent mental uncertainties, leading to inconsistent and unreliable predictions.
To address these limitations, this paper introduces ProMind-LLM.
We investigate an innovative approach integrating objective behavior data as complementary information alongside subjective mental records for robust mental health risk assessment.
Specifically, ProMind-LLM incorporates a comprehensive pipeline that includes domain-specific pretraining to tailor the LLM for mental health contexts, a self-refine mechanism to optimize the processing of numerical behavioral data, and causal chain-of-thought reasoning to enhance the reliability and interpretability of its predictions.
Evaluations of two real-world datasets, PMData and Globem, demonstrate the effectiveness of our proposed methods, achieving substantial improvements over general LLMs.
We anticipate that ProMind-LLM will pave the way for more dependable, interpretable, and scalable mental health case solutions.
\end{abstract}

\addtocontents{toc}{\protect\setcounter{tocdepth}{0}}

\section{Introduction}
\label{sec:intro}

\begin{figure}[t]
  \begin{center}
  \includegraphics[width=0.4\textwidth]{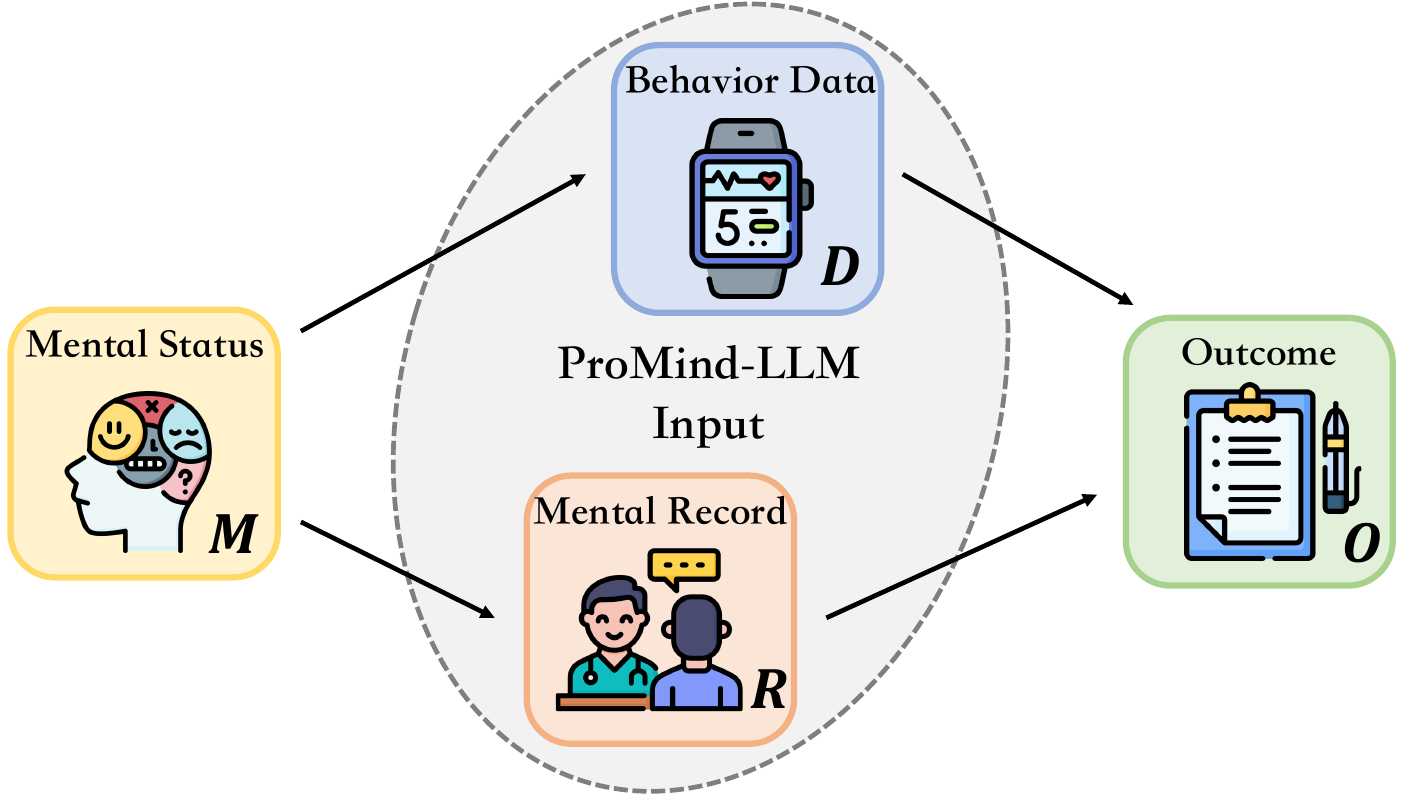}
  \caption{Causal relations among mental status, behavior data, and mental records. In ProMind-LLM, we utilize objective behavior data and subjective mental records to achieve robust mental risk prediction.}\label{fig:intro-causal}
  \vspace{-5mm}
  \end{center}
\end{figure}

In fact, mental health disorders have the highest prevalence rates compared to many other major health conditions.
According to WHO, approximately 25\% of people globally experience a mental health issue~\cite{world2019mental}.
Moreover, the COVID-19 pandemic and other global crises have exacerbated this issue, leading to a 25.6\% increase in anxiety and a 27.6\% rise in depression since 2020~\cite{mahmud2023global}.
The widespread nature of mental disorders underscores the need for developing automated detection tools to tackle this public health challenge~\cite{vos2020global}.

Currently, large language models (LLMs) have demonstrated remarkable success across a variety of domains~\cite{thirunavukarasu2023large, lv2024prollama, zhang2024empowering}.
Leveraging their advanced generalization and reasoning abilities, researchers have demonstrated their potential applications in mental health care~\cite{qiu2023smile, gabriel2024can, ji2024mindguard, ji2025transforming}.
LLMs can analyze user-generated data that reflects mental states, extracting contextual information to uncover subtle indicators of mental health disorders~\cite{hua2024large}.
For instance, prior works approved the effectiveness of using social media posts~\cite{kang2024cure, kumar2024mental} or conversational records~\cite{singh2024deciphering} to assess the presence of mental health issues.
Despite these promising results, these approaches are limited by their exclusive reliance on subjective textual data.
Such data can be influenced by transient factors, including the user's current mood or willingness to disclose accurate information, which undermines the reliability of predictions~\cite{who2022stigma}.
To address this limitation, incorporating objective factors, such as behavioral sensor data (\eg heart rate, sleep quality), offers a promising avenue to complement subjective data and enhance the accuracy of mental health predictions~\cite{patsali2020university, clement2021sleep}.
Recent studies have integrated other modalities such as speech~\cite{zhang2024llms} and video~\cite{singh2024deciphering} for mental disorder detection and achieved improvements compared to using subjective data only.
Nevertheless, the combined use of subjective records and objective behavior data for LLM-based mental health assessment remains largely underexplored.

Developing an LLM approach that effectively combines objective behavior data with subjective mental records poses several challenges. 
First, given the potential uncertainties in subjective mental records, LLMs must enhance their capacity to navigate the variability embedded within such data effectively.
%against the inherent variability in such data.
Second, most behavior data, such as heart rate, sleep quality, and exercise readings, are represented as lengthy numerical sequences~\cite{kim2024health, wu2024mindshift}.
Although these sequences can be formatted as text for LLM inputs, their sheer length and numeric nature can impede the model's ability for accurate interpretation~\cite{requeima2024llm, jin2024position}.
Third, enabling the LLM to fully utilize multimodal data, especially the causal relations~\cite{oftedal2019associations} between objective behavior data and subjective mental records, to further improve the outcome robustness requires a specific design.

To address these issues, we introduce ProMind-LLM, which builds upon the causal relations among mental status, mental records, and behavior data (\fig\ref{fig:intro-causal}), to deliver robust mental health risk assessment.
For the first challenge (\sec\ref{sec:training}), we construct a comprehensive mental health corpus and adopt continuous pertaining (PT)~\cite{gururangan2020don} to enhance the base model's understanding of mental health concepts, a widely used strategy in the development of prior mental health LLMs~\cite{zhai2024chinese, ji2023domain}.
Following PT, we implement counterfactual learning-based supervised finetuning (SFT), which generates misleading information in users' mental records to challenge and improve the LLM's resilience against uncertainties in subjective data.
ProMind-LLM addresses the second issue (\sec\ref{sec:self-refine}) by proposing a self-refine mechanism to format the behavior data, allowing the LLM to iteratively update and optimize the format based on its own feedback.
This approach not only shortens behavior inputs while preserving key features relevant to mental status but also enhances the model's ability to interpret behavior insights.
For the third challenge (\sec\ref{sec:causal-cot}), ProMind-LLM initially employs chain-of-thought (CoT)~\cite{wei2022chain} reasoning, integrating the causal relations between mental status and behavior.
This allows the model to analyze the mental records and behavior data individually and account for their mutual influence, rendering a more robust mental risk prediction.
Additionally, ProMind-LLM then utilizes counterfactual reasoning to further refine its judgment.
By exploring ``what-if'' scenarios-\eg, ``\textit{What if the user's eating habit were normal? Would this absence change the outcome?}''-the model evaluates alternative outcomes based on hypothetical changes to the input data~\cite{chen2022disco}.
This process enables the LLM to identify true indicators of mental risks, and ultimately confirm the outcome's reliability.
Together, these reasoning steps form the causal CoT scheme in ProMind-LLM.

To sum up, we make the following contributions in this paper:
\begin{itemize}
    \item To the best of our knowledge, ProMind-LLM is the first study to apply LLMs for mental health risk assessment by using both objective behavior data and subjective mental records, addressing the limitations of relying solely on subjective textual inputs.
    \item ProMind-LLM incorporates a comprehensive pipeline comprising domain-specific training, behavior data preprocessing via a self-refine scheme, and causal CoT reasoning for robust and interpretable predictions.
    \item Extensive experiments verify the effectiveness of ProMind-LLM over general LLMs, achieving improved outcome accuracy. This novel approach paves the way for realizing ubiquitous proactive mental health care.
\end{itemize}

\section{Related Work}
\label{sec:related-work}

\subsection{LLMs for Mental Health}
Leveraging NLP technologies for the early detection and intervention of mental health issues stands as a valuable endeavor~\cite{wang2024mentalmanip}.
With the advent of LLMs, researchers have recognized the potential of these tools to facilitate mental health care~\cite{lamichhane2023evaluation}.
Recent studies have evaluated the performance of state-of-the-art (SOTA) LLMs on various mental health tasks, including the detection of depression and anxiety~\cite{xu2024mental, yang2023towards} and stress level prediction~\cite{kim2024health}, exhibiting promising yet limited performance due to a lack of domain-specific knowledge.
To address this limitation, efforts have been made to adapt general-purpose LLMs into domain-specific models tailored for mental health applications.
Mental-BERT~\cite{ji2021mentalbert, ji2023domain} applied domain-specific pretraining to enhance its effectiveness in mental health tasks.
Similarly, Mental-Alpaca~\cite{xu2024mental} and MentalLlama~\cite{yang2024mentallama} employed instruction fine-tuning to improve corresponding reasoning capabilities.
In addition, Chinese MentalBERT~\cite{zhai2024chinese} is the first LLM that focuses on mental disorder detection for Chinese social media.
Beyond developing specialized models, researchers also emphasize the need for comprehensive benchmarks to support the advancement of mental health LLMs~\cite{singh2024deciphering, gabriel2024can, wang2024mentalmanip}.
Despite these advancements, most existing efforts primarily rely on subjective textual data.
ProMind-LLM takes a step forward by integrating objective behavioral data with subjective mental records, offering a more precise approach to proactive mental health care.

\subsection{LLMs for Numerical Data Understanding}

Despite being primarily used in the field of NLP, LLMs have shown remarkable potential in handling and reasoning about numerical data~\cite{liu2023large}.
One notable application is the classification of human activities using data from IMUs~\cite{ji2024hargpt, civitarese2024large}.
Similarly, LLMs have been used to analyze user physiological data, such as sleep quality, calorie intake, and heart rate, to generate personalized and professional health recommendations~\cite{yang2024drhouse, kim2024health, cosentino2024towards}.
These applications underscore the models' ability to interpret and apply numerical data in real-world contexts.
In addition to data interpretation, LLMs exhibit capabilities in time series forecasting~\cite{yang2024generative}.
These findings highlight the promise of LLMs for numerical data understanding.

\section{Methodology}
\label{sec:method}
\begin{figure*}[t]
  \begin{center}
  \includegraphics[width=0.9\textwidth]{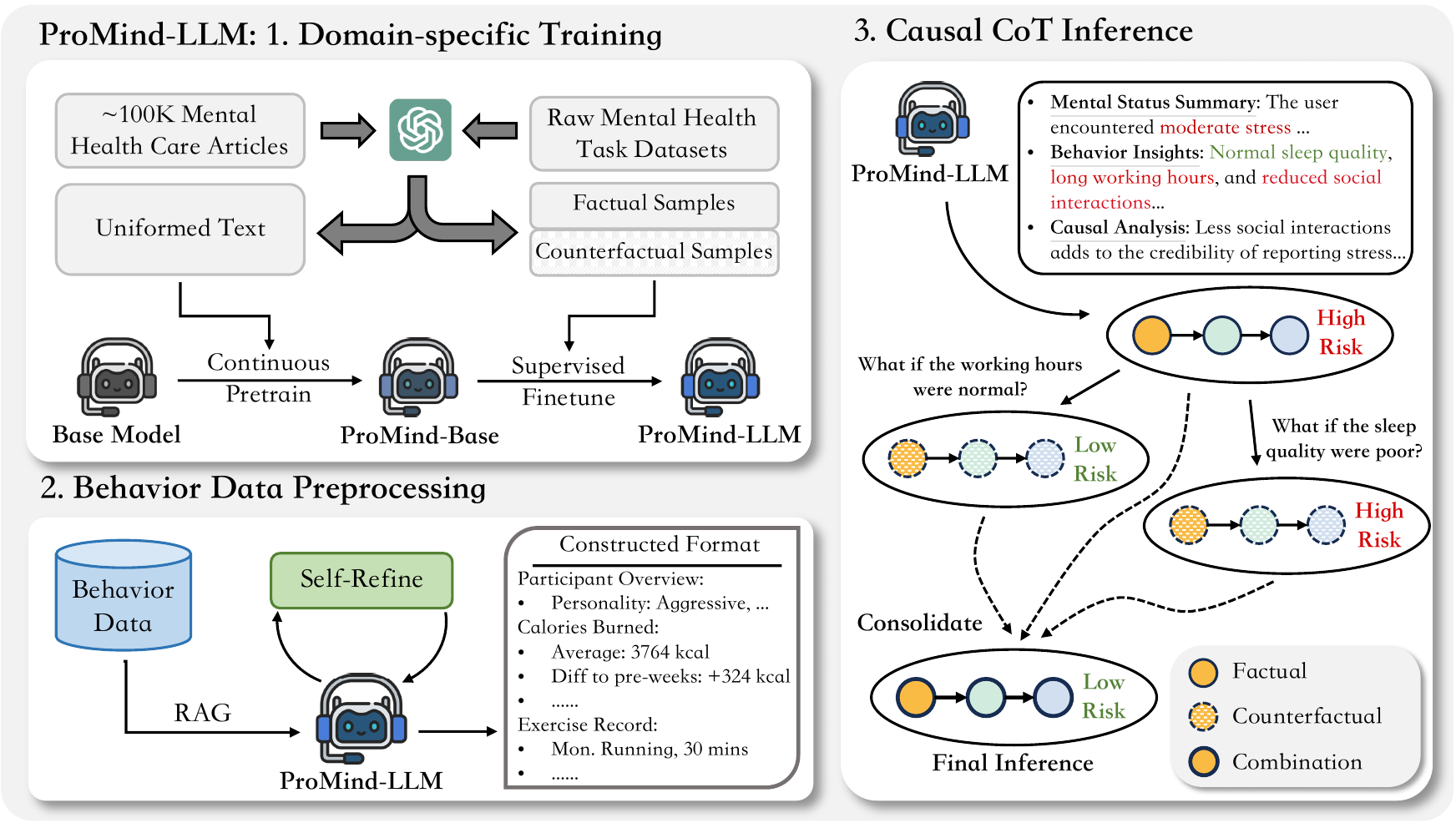}
  \caption{ProMind-LLM framework. The design consists of three components: (1) Construct professional mental health LLM using domain-specific training; (2) Enhance behavior understanding through self-refine mechanism; (3) Improve outcome robustness with causal CoT.}\label{fig:sys-framework}
  \end{center}
  \vspace{-3mm}
\end{figure*}

Our proposed method, as illustrated in \fig\ref{fig:sys-framework}, comprises three key components: domain-specific training, self-refine-based behavior data preprocessing, and causal CoT reasoning.
Together, these components enable robust mental risk prediction by integrating objective behavior data with subjective mental records.
Below, we provide a detailed explanation of each design.

\subsection{Domain-specific Training}
\label{sec:training}

\subsubsection{Continuous PT}
\label{sec:pt}

Continuous PT, also referred to as domain-adaptive PT, has been shown to significantly enhance performance, particularly for specialized downstream tasks~\cite{cui2021pre, gururangan2020don}.
To equip the base LLM—originally trained on a broad general corpus—with essential mental health domain knowledge, we adopt this approach by performing continuous PT on an extensive, mental health-specific corpus.

Unlike previous mental health-specific LLMs that predominantly rely on social media content~\cite{zhai2024chinese, zhai2024mentalglm}, we take a more comprehensive approach by curating approximately 100K professional mental health articles from the Web of Science, as suggested in~\cite{zhang2024empowering}.
These articles are selected using key terms outlined in \apx\ref{apx:pt-set}, covering topics like depression, anxiety, substance abuse, \etc.
The raw materials are then converted from PDF into structured TXT formats using MinerU~\cite{wang2024mineruopensourcesolutionprecise, he2024opendatalab}.
Further data cleansing~\cite{dubey2024llama} is implemented to correct the spelling errors and standardize format, such as removing redundant references, \etc.
The final PT corpus contains about 80 million tokens, preparing the base LLM with mental health expertise.
For details of the PT settings, please refer to \apx\ref{apx:pt-set}.

\subsubsection{Counterfactual-based SFT}
\label{sec:sft}
While continuous PT equips the base model with domain-specific expertise, SFT is an essential subsequent step, which refines the model's ability to effectively apply its knowledge in reasoning tasks with prompt instructions~\cite{zhang2023instruction}.
We take advantage of the public mental health-related datasets, including IMHI~\cite{yang2024mentallama}, CPsyCoun~\cite{zhang2024cpsycoun}, ANGST~\cite{hengle2024still}, and Depression Reddit~\cite{pirina2018identifying}, resulting in 112602 pieces of data.
These datasets provide high-quality question-answering pairs designed to analyze symptoms and infer outcomes based on the provided mental health records, equipping ProMind-LLM with greater precision and nuance to handle various mental health situations.
Please refer to \apx\ref{apx:sft-set} for statistical information on these datasets.

Counterfactual learning is proposed to further address the first challenge, which is to enhance the LLM's robustness against the uncertainties of subjective mental records.
This approach generates alternative scenarios where users' records might present misleading information about their mental states.
For instance, given an input SFT pair $\langle\bm{R}, \bm{O}\rangle$, where $\bm{R}$ denotes the mental record, and $\bm{O}$ presents the outcome analysis, we generate a counterfactual sample labeled $l$ within categories like ``personality traits'', ``stigma'', or ``lack of awareness''.
These elements are pivotal in distorting an individual to introduce biases or inaccuracies in self-reported data~\cite{who2022stigma}.
Finally, we generate two counterfactual parts for each pair and combine them into one single dataset to support counterfactual augmented SFT.
This augmentation enhances the model to recognize potential distortions, thereby improving its reliability in analyzing subjective mental health records.
The SFT settings can be found in \apx\ref{apx:sft-set}.
In addition, the prompt and the counterfactual sample are given in \apx\ref{apx:cf-aug}.

\subsection{Behavior Data Preprocessing}
\label{sec:self-refine}

Addressing the second challenge necessitates transforming numerical behavior data into a format that the LLM can efficiently process and comprehend.
Inputting lengthy sequences of raw behavior data directly into the LLM poses problems, including the risk of sequence truncation and memory constraints~\cite{gruver2024large}, as well as the LLM's inherent challenges in processing the digital data~\cite{chen2024see}.

To fully leverage ProMind-LLM's capacity for interpreting behavioral data, we propose a self-refine mechanism~\cite{madaan2024self}.
As depicted in \fig\ref{fig:sys-framework}, ProMind-LLM first retrieves raw behavior data using RAG~\cite{lewis2020retrieval}, evaluating the data's presentation format for redundancy and ease of understanding.
Based on this evaluation, the LLM provides an improved data version through self-refine.
This mechanism allows the LLM to update and optimize the data format autonomously based on its own feedback, thereby improving its ability to interpret behavior insights accurately.

\tab\ref{tab:self-refine} validates the effectiveness of our proposed method.
For instance, we use perplexity~\cite{alon2023detecting} to measure the LLM's familiarity with the behavior data format.
Compared with existing methods~\cite{ji2024hargpt, kim2024health, xu2024penetrative}, our approach not only reduces the input token length but also markedly enhances the LLM's proficiency in interpreting user behavior data.

We provide the corresponding pseudo-code in \apx\ref{apx:method-detail} for reference.

\subsection{Causal CoT}
\label{sec:causal-cot}

Given the processed behavioral data with users' mental records, ProMind-LLM is designed to assess the potential mental health risks.
Achieving accurate predictions demands a profound understanding of the data, coupled with a rigorous reasoning process.
Motivated by the intrinsic causal relations between behavioral data and mental records, we implement causal reasoning into the widely adopted CoT framework, resulting in causal CoT for ProMind-LLM.
This integration aims to further mitigate the impact of uncertainties in subjective data, enhancing the model's predictive accuracy.

\begin{table}[t]
\centering
\footnotesize
\caption{Assessment of the input format for behavioral data regarding token numbers and perplexity.}
\label{tab:self-refine}
% \resizebox{0.75\columnwidth}{!}{%
\small
\begin{tabular}{@{}ccc@{}}
\toprule
Method         & Tokens $\downarrow$ & Perplexity $\downarrow$ \\ \midrule
HARGPT~\cite{ji2024hargpt}         & $2^{11}$      & 15.63         \\
Health-LLM~\cite{kim2024health}     & $2^{12}$       & 17.22          \\
Penetrative AI~\cite{xu2024penetrative} & $2^{11}$       & 6.18          \\ \midrule
Ours           & $\mathbf{2^{9}}$      & \textbf{4.31}          \\ \bottomrule
\end{tabular}%
% }
% \vspace{-5mm}
\end{table}

As shown in \fig\ref{fig:sys-framework}, the causal CoT comprises three components, CoT on factual samples, CoT on counterfactual samples, and a combination analysis of the two.
First, we prompt ProMind-LLM under the CoT framework to conduct a preliminary analysis of mental records ($\bm{R}$) and behavioral data ($\bm{D}$) individually.
It then leverages the causal relation between the two modalities to evaluate their mutual effects.
For example, if both modality data present high-risk indicators, such as ``moderate stress'' and ``reduced social interactions'', the model may identify a possible correlation between these factors.
The greater the number of associated indicators, the higher the likelihood of potential mental risks for the user.
We use the following formula to represent the first step:
\begin{equation}
\begin{split}
    \bm{A} = \{(d_{i}, r_{j}) \mid d_{i} \in \bm{D}, r_{j} \in \bm{R},
    P(r_{j} | d_{i}) > \tau\},
\end{split}
\end{equation}
where $\bm{A}$ denotes the factual analysis results, $d_{i}$ and $r_{j}$ represent the risk indicators in behavior data and mental records respectively, $P$ indicates the causal analysis by ProMind-LLM, and$(d_{i}, r_{j})$ is the pair found with a possible causal relation.

To validate and refine the results of the initial step, ProMind-LLM utilizes counterfactual reasoning.
This involves exploring ``what-if'' questions to strengthen the causal understanding of certain pairs, reanalyze weak correlations, and address pairs that might be overlooked due to uncertainties in mental records.
For example, by asking: ``\textit{What if the user's working hours were reduced, would the user's stress be better?}''.
If the model believes that stress can be alleviated to a certain extent, as the user may have more time to restart the social interactions, then the social interactions and stress pair built in the first step would be less critical.
By disentangling these causal influences, the model achieves a more nuanced understanding of the user's mental health dynamics.
We summarize the counterfactual reasoning as follows:
\begin{equation}
\begin{split}
    \bm{A}_{c}=\{(d_{i}, r_{j})|d_{i} \in \bm{D}, r_{j} \in \bm{R}, \\P(r_{j} | d_{i}, c) > \tau\},
\end{split}
\end{equation}
where $\bm{A}_{c}$ indicates the results in counterfactual reasoning, and $c$ is the counterfactual reasoning.

The final step of causal CoT is the integration of the insights from both factual and counterfactual analysis to generate its final predictions: $(\bm{A}, \bm{A}_{c}) \rightarrow \bm{G}$, where $\bm{G}$ is the user's potential mental health risk.
Please refer to \apx\ref{apx:mental-risk-pred} for detailed examples.

\section{Experimental Setup}
\label{sec:exp-set}

\subsection{Task Definition}
\label{sec:task-def}
To evaluate the effectiveness of our methods and the performance of ProMind-LLM in predicting mental health risks, we aim to address a binary classification task.
Each individual contains long-term recordings of both behavioral data ($\bm{D}$) and self-reported mental records ($\bm{R}$).
These records are aggregated weekly and represented as $\langle\bm{D}_{i}, \bm{R}_{i}\rangle$.

Based on these inputs, LLMs need to predict the potential mental health risk label $\bm{G}_i$, categorized as follows:
\begin{enumerate}[leftmargin=*]
    \item $\bm{G}_i=0$: The individual shows no significant signs of mental health issues, or may have minor issues that do not require immediate psychological intervention.
    \item $\bm{G}_i=1$: The individual exhibits strong indicators of mental health issues and requires further professional treatment or closer monitoring.
\end{enumerate}

\subsection{Dataset Description}
\label{sec:dataset-des}

We conduct experiments on two open-sourced datasets, PMData~\cite{thambawita2020pmdata} and Globem~\cite{xu2022globem}, which provide behavioral data and self-reported mental health records but lack predefined mental health risk labels ($\bm{G}_i$), necessitating manual annotation.
Based on the professional criteria~\cite{morgan2018systematic, guha2014diagnostic}, we assess existing or potential mental health risks for each recording.
To mitigate the potential false positives and negatives, we cooperate with licensed psychological experts to review the initial assessments and refine the labels by considering behavior patterns, and inconsistencies in self-reported records, thereby enhancing the reliability of the labeling process.

The details of each dataset are given below\footnote{All data usage strictly adheres to the Data Use Agreements of PMData (CC BY 4.0) and Globem (PhysioNet Credentialed Health Data License 1.5.0).}:

\head{PMData}
The dataset comprises 16 participants monitored over 5 months using Fitbit for objective biometrics and activity data, Google Forms for demographics, food, drinking, and weight data, and the PMSys for self-reported measures such as fatigue, mood, stress, etc.
The data collected from Fitbit and Google Forms constitute the participants' behavior data ($\bm{D}_{i}$), while the PMSys measures represent their self-reported mental records ($\bm{R}_{i}$).
All participants are used for evaluation.
A small fraction of 9.8\% cases are identified with potential mental health issues requiring additional support.

\head{Globem}
The Globem dataset encompasses four years of passive sensing data from 497 participants.
Behavioral data, including sleep patterns, location, physical activity, and phone usage, are collected using wearable sensors (Fitbit Flex2 and Inspire 2) and are denoted as $\bm{D}_{i}$.
Survey data, such as PHQ-4~\cite{kroenke2009ultra} (mental health, anxiety, and depression), PSS-4~\cite{cohen1983global} (stress level), and PANAS~\cite{watson1988development} (positive and negative affect), provide self-reported mental health records ($\bm{R}_{i}$).
For efficient testing and cost reduction, 25\% of the participants are randomly chosen as the test set.
Within this group, 23.2\% are identified to have potential risks.

\section{Results and Analysis}
\label{sec:res-ana}

\subsection{Evaluation Design}
\label{sec:eva-design}

Evaluation is designed to answer the following key research questions:
1): How effective is the proposed two-stage domain-specific training in enhancing the model's ability to assess mental health risks? (\sec\ref{sec:effect-train})
2): Do the self-refine mechanism and causal CoT improve assessment accuracy, and to what extent? (\sec\ref{sec:ablation}).
3): By employing counterfactual data augmentation in SFT, does ProMind-LLM demonstrate improved resilience against uncertainties in mental records? (\sec\ref{sec:ablation})
4): To what extent does behavioral data serve as a complementary modality to subjective mental records in enhancing mental health risk assessment? (\sec\ref{sec:ablation-behavior})
5): Does the model's analysis align from end to end, and how consistent are the evidence and outcomes of the analysis? (\sec\ref{sec:consistency-measure})

\subsection{Baselines and Deployment Details}
We select several baseline models for performance comparison, including three SOTA commercial LLMs (GPT-4o, GPT-3.5, and Claude-3.5) and two leading open-sourced LLMs (LLaMA3-Chat-70B and Qwen2-Chat-72B).
Additionally, two domain-specific models, Mental-Alpaca~\cite{xu2024mental} and MentalLlama~\cite{yang2024mentallama}, are also evaluated but excluded due to inadequate outputs (see \apx\ref{apx:fail-examples}).

To develop ProMind-LLM, we select two base LLMs: LLaMA3-base-8B and InternLM2-base-7B, prioritizing the future deployment of this application on the edge devices, a strategic decision aimed at minimizing potential privacy risks associated with data transfers.
For comparisons, we also include the vanilla versions of these models, serving as our open-sourced baselines.
Among these, ProMind-LLM developed from InternLM2-base-7B demonstrates a better performance, representing the ProMind-LLM in the rest of the paper if not otherwise indicated.

\begin{table*}[t]
\centering
\caption{Results for mental health issue binary classification. All methods here use both self-refine and causal CoT for mental risk prediction. {\rm (The best and second results are highlighted in \textbf{bold} and \underline{underlined}, respectively.)}}
\label{tab:overall-results}
\begin{threeparttable}
\resizebox{0.95\textwidth}{!}{
\begin{tabular}{@{}ccccccccccc@{}}
\toprule
\multirow{2}{*}{\textbf{Category}}     & \multirow{2}{*}{\textbf{Model}}            & \multirow{2}{*}{\textbf{Type}} & \multicolumn{4}{c}{\textbf{PMData}}                                        & \multicolumn{4}{c}{\textbf{Globem}}                                        \\ \cmidrule(l){4-11} 
                              &                                   &                       & \textbf{Accuracy}       & \textbf{Precision}      & \textbf{Recall}         & \textbf{F1}             & \textbf{Accuracy}       & \textbf{Precision}      & \textbf{Recall}         & \textbf{F1}             \\ \midrule
\multirow{3}{*}{Comercial}    & GPT-4o\tnote{1}                            & \textbackslash             & \textbf{0.956} & \textbf{0.781} & \underline{0.821}    & \textbf{0.800} & \textbf{0.867} & \textbf{0.663} & \underline{0.955}    & \textbf{0.783} \\
                              & GPT-3.5\tnote{1}                           & \textbackslash             & 0.858          & 0.385          & 0.714          & 0.500          & 0.747          & 0.505          & 0.918          & 0.651          \\
                              & Claude-3.5\tnote{1}                        & \textbackslash             & 0.923          & 0.786          & 0.314          & 0.449          & 0.789          & 0.552          & 0.951          & 0.699          \\ \midrule
\multirow{4}{*}{Open-sourced} & LLaMA3-chat-70B                   & \textbackslash             & 0.795          & 0.297          & 0.771          & 0.429          & 0.807          & 0.579          & 0.833          & 0.683          \\
                              & QWen2-chat-72B                    & \textbackslash             & 0.918          & 0.674          & 0.547          & 0.604          & 0.819          & 0.608          & 0.783          & 0.684          \\
                              & LLaMA3-chat-8B                    & \textbackslash             & 0.613          & 0.174          & 0.641          & 0.275          & 0.531          & 0.325          & 0.818          & 0.465          \\
                              & InternLM2-chat-7B                 & \textbackslash             & 0.634          & 0.191          & 0.828          & 0.310          & 0.367          & 0.289          & \textbf{1.000} & 0.449          \\ \midrule
\multirow{4}{*}{Ours}         & \multirow{2}{*}{LLaMA3-base-8B}   & SFT                   & 0.625          & 0.213          & \textbf{0.846} & 0.340          & 0.636          & 0.401          & 0.939          & 0.564          \\
                              &                                   & PT+SFT                & 0.883          & 0.492          & 0.769          & 0.601          & 0.754          & 0.504          & 0.901          & 0.649          \\ \cmidrule(l){2-11} 
                              & \multirow{2}{*}{InterLM2-base-7B} & SFT                   & 0.867          & 0.443          & 0.633          & 0.521          & 0.726          & 0.481          & 0.836          & 0.611          \\
                              &                                   & PT+SFT                & \underline{0.938}    & \underline{0.765}    & 0.667          & \underline{0.712}    & \underline{0.859}    & \underline{0.646}    & 0.939          & \underline{0.765}    \\ \bottomrule
\end{tabular}%
}
\begin{tablenotes}
    \footnotesize
    \item[1] GPT-4o API is ``gpt-4o-2024-05-13''. GPT-3.5 API is ``gpt-3.5-turbo''. Claude-3.5 API is ``claude-3-5-sonnet-20240620''.
\end{tablenotes}
\end{threeparttable}
\end{table*}

\begin{table*}[t]
\centering
\small
\caption{Ablation study on self-refine and causal CoT.  {\rm (The best and second results of each model are highlighted in \textbf{bold} and \underline{underlined}, respectively.)}}
\label{tab:ablation}
\begin{threeparttable}
\resizebox{0.95\textwidth}{!}{
\begin{tabular}{@{}ccccccccccc@{}}
\toprule
\multirow{2}{*}{\textbf{Model}}                              & \multicolumn{2}{c}{\textbf{Method}}                                & \multicolumn{4}{c}{\textbf{PMData}}                                        & \multicolumn{4}{c}{\textbf{Globem}}                                        \\ \cmidrule(l){2-11} 
                                                    & \textbf{Self-Refine}             & \textbf{Causal CoT}                  & \textbf{Accuracy}       & \textbf{Precision}      & \textbf{Recall}         & \textbf{F1}             & \textbf{Accuracy}       & \textbf{Precision}      & \textbf{Recall}         & \textbf{F1}             \\ \midrule
\multirow{4}{*}{GPT-4o}                             & \XSolidBrush & \XSolidBrush   & 0.938          & 0.734          & 0.726          & 0.729          & 0.803          & 0.575          & 0.818          & 0.675          \\
                                                    & \Checkmark   & \XSolidBrush & \underline{0.952}    & \underline{0.737}    & \underline{0.800}    & \underline{0.767}    & 0.814          & 0.586          & \underline{0.951}    & \underline{0.725}    \\
                                                    & \XSolidBrush & \Checkmark   & 0.933          & 0.687          & 0.769          & 0.726          & \underline{0.829}    & \underline{0.615}    & 0.848          & 0.713          \\
                                                    & \Checkmark   & \Checkmark   & \textbf{0.956} & \textbf{0.781} & \textbf{0.821} & \textbf{0.800} & \textbf{0.867} & \textbf{0.663} & \textbf{0.955} & \textbf{0.783} \\ \midrule
\multicolumn{1}{l}{\multirow{4}{*}{Qwen2-chat-72B}} & \XSolidBrush & \XSolidBrush & 0.837          & 0.382          & \textbf{0.693} & 0.495          & 0.761          & 0.516          & 0.742          & 0.609          \\
\multicolumn{1}{l}{}                                & \Checkmark   & \XSolidBrush & \underline{0.903}    & \underline{0.514}    & 0.514          & \underline{0.514}    & 0.769          & 0.523          & \textbf{0.879} & \underline{0.655}    \\
\multicolumn{1}{l}{}                                & \XSolidBrush & \Checkmark   & 0.859          & 0.423          & \underline{0.635}    & 0.507          & \underline{0.781}    & \underline{0.541}    & \underline{0.798}    & 0.645          \\
\multicolumn{1}{l}{}                                & \Checkmark   & \Checkmark   & \textbf{0.918} & \textbf{0.674} & 0.547          & \textbf{0.604} & \textbf{0.819} & \textbf{0.608} & 0.783          & \textbf{0.684} \\ \midrule
\multirow{4}{*}{ProMind-LLM}                        & \XSolidBrush & \XSolidBrush & 0.845          & 0.369          & 0.504          & 0.426          & 0.734          & 0.475          & 0.641          & 0.546          \\
                                                    & \Checkmark   & \XSolidBrush & \textbf{0.940} & \textbf{0.792} & \underline{0.543}    & \underline{0.644}    & \underline{0.844}    & \underline{0.640}    & \underline{0.902}    & \underline{0.748}    \\
                                                    & \XSolidBrush & \Checkmark   & 0.849          & 0.376          & 0.479          & 0.422          & 0.766          & 0.525          & 0.672          & 0.589          \\
                                                    & \Checkmark   & \Checkmark   & \underline{0.938}    & \underline{0.765}    & \textbf{0.667} & \textbf{0.712} & \textbf{0.859} & \textbf{0.646} & \textbf{0.939} & \textbf{0.765} \\ \bottomrule
\end{tabular}
}
\end{threeparttable}
\end{table*}

\subsection{Effectiveness of Training Strategy}
\label{sec:effect-train}

We evaluate the performance of ProMind-LLM against baseline models on PMData and Globem datasets.
Additionally, we compare the outcomes of applying SFT directly to two base LLMs versus a two-stage process involving PT plus SFT.
The results are summarized in \tab\ref{tab:overall-results}.
All methods implement both self-refine mechanism and causal CoT reasoning.
We exclude the base models due to their lack of instruction following capabilities.

GPT-4o outperforms other baseline models across both datasets.
Notably, GPT-4o achieves a recall rate of 0.821 on PMData and 0.955 on Globem.
Among open-source models, Qwen2-chat-72B achieves the highest F1 scores, 0.604 and 0.684, on PMdata and Globem, respectively.

The results further highlight a significant improvement when using the combined PT and SFT approach.
For instance, on PMData dataset, InternLM2-7B shows substantial gains, with precision and recall increasing by 42.1\% and 5.1\%, respectively, and an overall F1 score improvement of 26.8\% compared to the model trained without PT.
This trend is consistently observed on the Globem dataset, where the combined PT and SFT strategy outperforms SFT alone, further validating the benefits of our proposed training framework.

Remarkably, ProMind-LLM achieves a performance ranking second only to GPT-4o in both datasets, even surpassing leading open-sourced LLMs that are ten times its size.
The results demonstrate that (i) ProMind-LLM not only integrates mental health knowledge into its base model but also matches the reasoning capabilities of SOTA large-scale commercial LLM, GPT-4o. (ii)
ProMind-LLM analyzes users' mental health conditions with performance that surpasses its counterparts possessing more than 10 times its parameters (LLaMA3-70B, etc.).

\subsection{Ablation Study}
\label{sec:ablation}
\head{Self-refine \& Causal CoT}

To verify the effectiveness of self-refine mechanism and causal CoT, we conduct an ablation study using GPT-4o and Qwen2-chat-72B as baseline models, with results presented in \tab\ref{tab:ablation}.
In this experiment, disabling the self-refine mechanism involves directly feeding the original behavior sequence data into the model, while disabling causal CoT indicates using CoT only for the final analysis.
Across all models, the combination of both strategies consistently yields the best performance.
For instance, ProMind-LLM achieves F1 score improvements of 40.2\% on PMData and 28.6\% on Globem compared to not using either method.
In contrast, GPT-4o demonstrates an average F1 improvement of only 11.3\%, likely due to its already strong generalization and reasoning capabilities.

Both self-refine and causal CoT reasoning independently enhance model performance, with self-refine yielding greater overall gains.
For GPT-4o, self-refine and causal CoT improve F1 scores by 5\% and 2\%, respectively, while Qwen2-chat-72B achieves gains of 3.4\% and 3.1\%.
ProMind-LLM benefits significantly from self-refine mechanism, with a 24.9\% F1 increase, likely due to challenges small-scale LLMs face in processing lengthy numerical sequences, which can hinder contextual understanding and reasoning performance.

\begin{figure}[t]
  \begin{center}
  \includegraphics[width=0.47\textwidth]{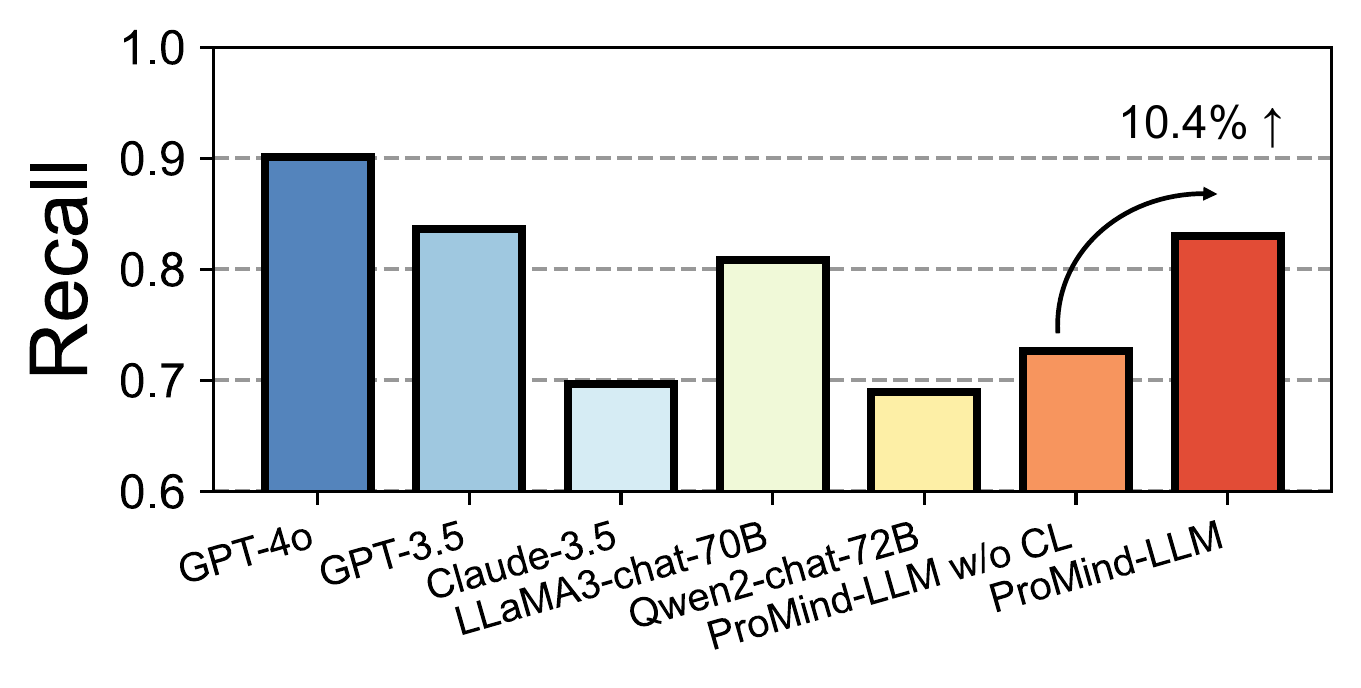}
  % \vspace{-5mm}
  \caption{Ablation study on counterfactual augmentations in SFT.}\label{fig:counterfactual-sft}
  \end{center}
\end{figure}

\head{Counterfactual-based SFT}

During the SFT stage, we use counterfactual data augmentation to introduce misleading information into users' subjective mental health records, enhancing the model's ability to handle subjective uncertainties.
As shown in \fig\ref{fig:counterfactual-sft}, we calculate the average recall rate using both datasets.
This technique improves ProMind-LLM's recall from 72.6\% to 83\%, outperforming open-sourced LLMs.

\subsection{Behavior Data for Complementary}
\label{sec:ablation-behavior}

\begin{figure}[t]
  \begin{center}
  \includegraphics[width=0.47\textwidth]{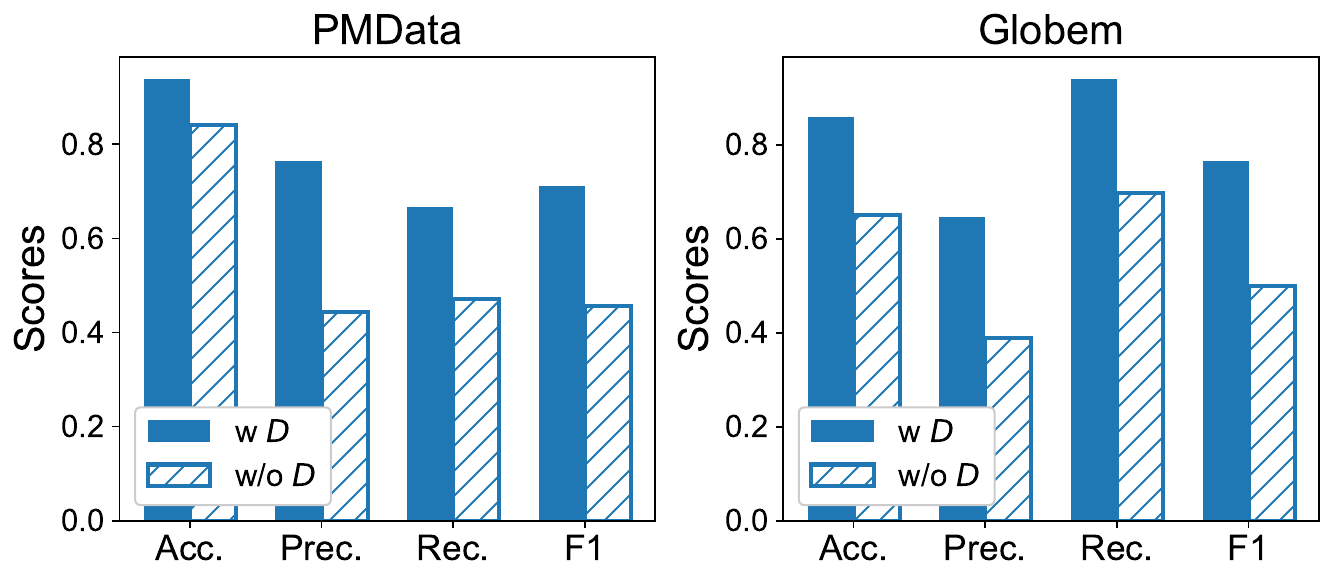}
  \vspace{-3mm}
  \caption{Ablation study on behavior data as complementary modality.}\label{fig:behavior-ablation}
  \end{center}
\end{figure}

To assess the influence of behavioral data as complementary modalities on outcomes, we perform an ablation study detailed in \fig\ref{fig:behavior-ablation}.
The findings underscore the importance of integrating objective behavioral data with subjective mental records for accurate analysis.

\subsection{Consistency Measurement}
\label{sec:consistency-measure}

\begin{figure}[t]
  \begin{minipage}{0.235\textwidth}
    \centering
    \includegraphics[width=0.95\textwidth]{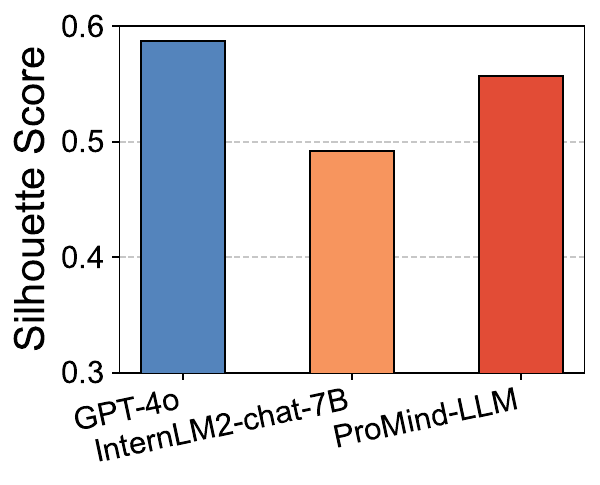}
    \vspace{-0.1in}
    \caption{Silhouette scores of the word embedding.}
    \label{fig:silhouette}
  \end{minipage}
  \hfill
  \begin{minipage}{0.235\textwidth}
    \centering
    \includegraphics[width=0.95\textwidth]{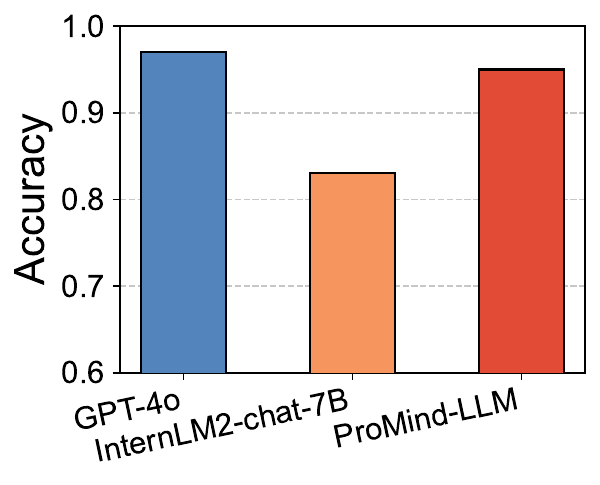}
    \vspace{-0.1in}
    \caption{Accuracy of consistency.}
    \label{fig:consis-acc}
  \end{minipage}
\end{figure}

Following the methodology in~\cite{yang2024mentallama}, we assess whether the analysis evidence supports the outcomes.
Using pre-trained BERT~\cite{devlin2018bert} for embedding extraction, we calculate the Silhouette score~\cite{shahapure2020cluster} to evaluate clustering quality, with higher scores indicating better performance.
As shown in \fig\ref{fig:silhouette}, ProMind-LLM, with domain-specific training, outperforms InternLM2-chat-7B, achieving a 0.065 improvement in the Silhouette score.
Additionally, we introduce a classification network following the embedding extraction and employ K-fold cross-validation to calculate the overall accuracy of classifying outcomes based on the evidence in the analysis, as presented in \fig\ref{fig:consis-acc}.
ProMind-LLM achieves an accuracy of 0.95, slightly lower than GPT-4o (0.97), which proves the consistency of the evidence and the outcome.
In contrast, InternLM2-chat-7B achieves an accuracy of 0.83, further validating the robustness of ProMind-LLM.

\section{Conclusion}

This paper presents ProMind-LLM, the first LLM-based approach to integrate both objective behavioral data and subjective mental records for mental risk analysis.
Our pipeline includes domain-specific training to equip the base model with mental health concepts, the self-refine mechanism for behavior data preprocessing to enhance numerical data comprehension, and causal CoT reasoning to deliver precise predictions by leveraging causal relations between mental records and behavioral data.
The experimental results validate the effectiveness of our methods, highlighting ProMind-LLM's ability to advance the research in this domain.
We believe ProMind-LLM marks a significant step toward achieving proactive mental health care.

\section*{Limitations}
While ProMind-LLM demonstrates strong performance on public datasets such as PMData and Globem, which encompass a four-year longitudinal study, its effectiveness in real-world, uncontrolled environments with diverse populations, in terms of age, gender, and race, remains insufficiently validated.
Additionally, due to resource constraints, our proposed training pipeline has been evaluated only on small-scale LLMs.
Moreover, although we have conducted a preliminary human evaluation assessing the readability of causal CoT reasoning steps and analyzing the causes of incorrect predictions (\sec\ref{apx:human-eval}), a more comprehensive human evaluation is required.
Specifically, further investigation into misclassified cases—such as underestimation of missed risks—is essential to ensure the system's reliability and safety, given its role as a mental health recommendation system.
This expanded evaluation will also provide deeper insights into the specific scenarios where ProMind-LLM is prone to errors, guiding future refinements to enhance its robustness and overall effectiveness.

\section*{Ethics Statement}
This research complies with ethical standards to ensure integrity and participant welfare. The publicly available datasets, PMData and Globem, were utilized under their Data Use Agreements, with all participant data anonymized to protect privacy. Original data collection adhered to informed consent protocols, ensuring ethical compliance. Licensed psychological experts validated mental health risk assessments, minimizing inaccuracies and maintaining sensitivity towards participants. Methodological transparency and reproducibility were upheld through detailed documentation of pretraining, finetuning, and counterfactual augmentation processes. To avoid biases, counterfactual learning techniques were employed, enhancing the model’s robustness against uncertainties in subjective mental records. Efforts were made to mitigate biases in mental health assessments by incorporating counterfactual learning and rigorous evaluation techniques to ensure fair, non-discriminatory analysis while avoiding stigmatization. This work reflects a commitment to advancing mental health care responsibly and with respect for human dignity.

\section*{Acknowledgment}
This research was funded in part by National Institutes of Health (NIH) under award \#P41EB028242.
We also acknowledge the support from the Heritage Medical Research Institute.

\newpage
\bibliography{custom}

\newpage
\appendix
\renewcommand{\contentsname}{Appendix}
\tableofcontents
\addtocontents{toc}{\protect\setcounter{tocdepth}{2}}

\section{Risk Assessment}
Although ProMind-LLM has been evaluated solely on public datasets, applying an LLM-based mental health risk assessment approach in real-world settings introduces potential ethical risks.
Inaccurate predictions may worsen users' conditions or lead to inappropriate recommendations.
To gradually mitigate this concern, ongoing validation, rigorous monitoring, and human oversight are critical to ensure the system's reliability and safety.

\section{Experimental Settings}

\subsection{Pretraining Settings}
\label{apx:pt-set}
In the PT section, we deploy Deepspeed ZeRO-2~\cite{rasley2020deepspeed} and flash-attention2~\cite{dao2023flashattention} to improve memory efficiency, set the global batch size to 96, and epoch to 1.
We use a learning rate warmup of 5 \% of the total training steps, followed by cosine annealing of the learning rate, the maximum learning rate during this period is $5 \times 10^{-5}$.
We use AdamW optimizer\cite{loshchilov2017decoupled} with a weight decay factor of $1 \times 10^{-2}$ and gradient clipping with the maximum grad norm of 1.0.
To use more tensor cores, we train with mixed precision, where computation is done within the bfloat16 datatype.
To mitigate the potential for catastrophic forgetting, we incorporate a diverse dataset consisting of 160M tokens sourced from the RefinedWeb dataset, which is subsequently mixed with an additional 80M tokens of domain-specific data of mental health.
The keywords for those domain-specific documents are listed in \tab\ref{tab:keywords}.
This approach is employed to enhance the model's robustness and maintain its proficiency across a broad spectrum of tasks.

\begin{table}[ht]
\centering
\caption{Summary of Corpus Keywords and Article Distribution}
\label{tab:keywords}
\resizebox{0.9\columnwidth}{!}{%
\begin{tabular}{@{}ccc@{}}
\toprule
\textbf{Category}           & \textbf{Key Words}           & \textbf{\# of Articles} \\ \midrule
\multirow{2}{*}{General}    & Mental health first aid      & 1809                    \\
                            & Mental health                & 7335                    \\ \midrule
\multirow{11}{*}{Disorders} & Depression                   & 9007                    \\
                            & Anxiety                      & 8356                    \\
                            & Bipolar                      & 8406                    \\
                            & Eating disorders             & 7707                    \\
                            & Stress management            & 7987                    \\
                            & Suicide                      & 7973                    \\
                            & Cognitive behavioral therapy & 9358                    \\
                            & Grief                        & 6086                    \\
                            & PTSD                         & 8808                    \\
                            & Schizophrenia                & 9014                    \\
                            & Substance abuse              & 9008                    \\ \midrule
Sum                         & -                            & 100854                  \\ \bottomrule
\end{tabular}
}
\end{table}

\begin{table}[ht]
\centering
\caption{Statistical information of SFT datasets}
\label{tab:sft-datasets}
\resizebox{0.95\columnwidth}{!}{%
\begin{tabular}{@{}cccc@{}}
\toprule
Dataset           & Category          & Topics                                                                                                   & Size   \\ \midrule
IMHI              & Social Media Post & \begin{tabular}[c]{@{}c@{}}depression, stress,\\ suicide, loneliness,\\ wellness dimensions\end{tabular} & 105792 \\ \midrule
ANGST             & Social Media Post & depression, anxiety                                                                                      & 2876   \\ \midrule
Depression Reddit & Social Media Post & depression                                                                                               & 800    \\ \midrule
CPsyCoun          & Conversation      & diverse                                                                                                  & 3134   \\ \bottomrule
\end{tabular}%
}
\end{table}

\subsection{Supervised Finetuning Settings}
\label{apx:sft-set}
\tab\ref{tab:sft-datasets} represents the statistical information of the datasets utilized in the SFT stage, including their categories, topics, and the corresponding data counts for each.

In the SFT stage, we follow the parameter Settings of the PT stage but reduce the maximum learning rate to 5e-6, one-tenth of PT, and set the training epoch to 2.

The PT and SFT are deployed using Llama-factory~\cite{zheng2024llamafactory} framework and performed with 8 NVIDIA A100-80G SXM4 GPUs.
The PT costs approximately 12 hours, and SFT costs around 2 hours.

\section{Methodological Details}
\label{apx:method-detail}

We provide the pseudo code for self-refine-based behavior data preproessing (\alg\ref{alg:self_refine}) in this section.

\begin{algorithm}[ht]
\caption{Self-Refine Mechanism in ProMind-LLM}
\label{alg:self_refine}
\begin{algorithmic}[1]
\REQUIRE Raw behavioral data $\bm{D}_{\text{raw}}$, base LLM $\mathcal{M}$, maximum refinement loops $k$
\ENSURE Refined behavioral data $\bm{D}_{\text{refined}}$

\STATE Initialize $\bm{D}_{\text{current}} \gets \bm{D}_{\text{raw}}$
\FOR{$i = 1$ to $k$}
    \STATE Evaluate redundancy and comprehensibility of $\bm{D}_{\text{current}}$ using $\mathcal{M}$
    \STATE Generate refined data $\bm{D}_{\text{new}}$ based on feedback from $\mathcal{M}$
    \STATE Update $\bm{D}_{\text{current}} \gets \bm{D}_{\text{new}}$
\ENDFOR
\STATE $\bm{D}_{\text{refined}} \gets \bm{D}_{\text{current}}$
\RETURN $\bm{D}_{\text{refined}}$
\end{algorithmic}
\end{algorithm}

\section{Generalization Analysis}
\label{apx:generalization-analysis}

\begin{figure}[ht]
  \begin{center}
  \includegraphics[width=0.8\columnwidth]{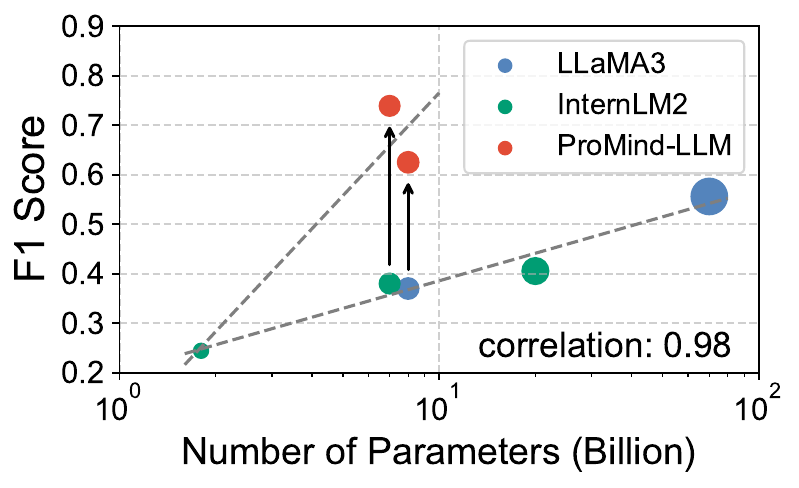}
  \vspace{-3mm}
  \caption{Scalability analysis on overall F1 score.}\label{fig:scaling}
  \end{center}
\end{figure}

We conduct a scalability analysis to demonstrate that our proposed method not only enhances LLM performance in mental health scenarios but also reduces model costs. \fig\ref{fig:scaling} illustrates the scaling law for the InternLM2 and LLaMA3 series, showing the predictable relationship between model size and F1 scores on the MHFA task.
By building ProMind-LLM on these models, we have been able to break through the existing scaling laws, achieving performance improvements of over 50\% and exceeding even the largest counterparts in the series.
This new scaling law trend suggests that ProMind-LLM offers substantial computational cost-effectiveness.

\begin{figure}[ht]
  \begin{center}
\includegraphics[width=0.8\columnwidth]{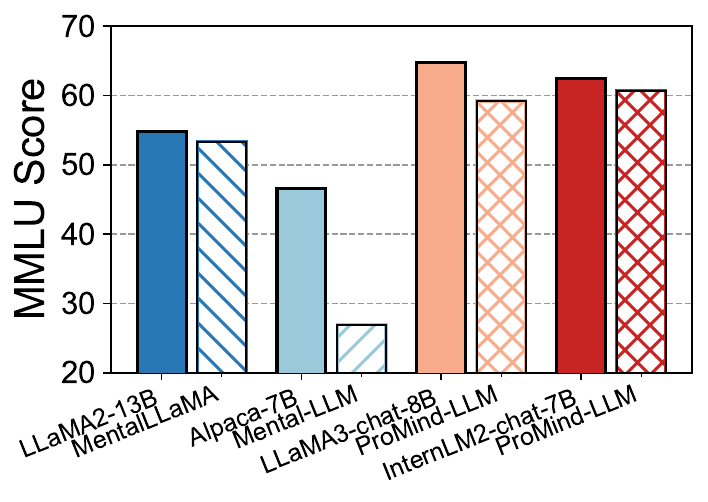}
\vspace{-3mm}
  \caption{Generalization test on different mental health LLMs and their base models.}\label{fig:mmlu}
  \end{center}
\end{figure}

Multi-task language understanding (MMLU)~\cite{hendrycks2020measuring} is a key metric for assessing the general capabilities of LLMs.
Domain-specific LLMs, like ProMind-LLM, often encounter performance decrease with domain-specific training; hence, we evaluate it using the MMLU benchmark, which tests LLMs across a variety of subjects and tasks, the result is shown in \fig\ref{fig:mmlu}.
ProMind-LLM demonstrates robust reasoning and language understanding, effectively reducing the risk of catastrophic forgetting in its domain.
In contrast, previous works such as MentalLlama~\cite{yang2024mentallama}, despite integrating LoRA~\cite{hu2021lora} for SFT, only managed to maintain an acceptable level of performance.
Meanwhile, Mental-LLM~\cite{xu2024mental} lost the general capabilities that are fundamental to the base LLM as it overfits the input-output training pairs.
Thus, these two models entirely fail on our tasks, we present their generated contents in \apx\ref{apx:fail-examples}.

\section{Human Evaluation}
\label{apx:human-eval}

To better understand the potential limitations of ProMind-LLM, we conduct a preliminary human expert evaluation focusing on two key aspects: the interpretability of causal CoT reasoning and the reason behind error cases.
The first evaluation examines whether the model’s reasoning process is logically coherent and aligned with domain expertise, while the second categorizes misclassified cases to identify common sources of error.
Details of the evaluation are presented below.

\begin{figure}[ht]
  \begin{center}
\includegraphics[width=0.8\columnwidth]{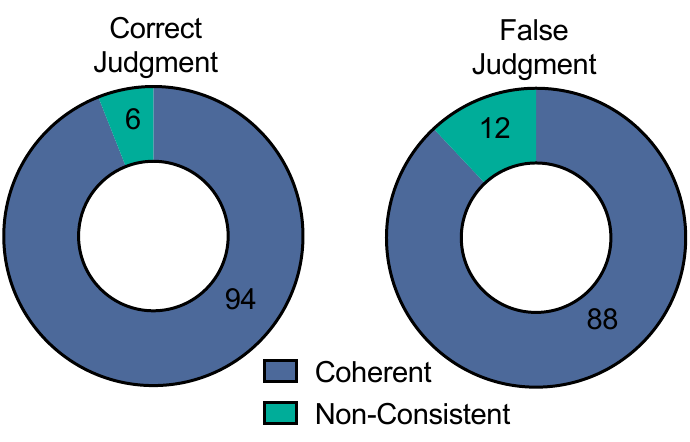}
\vspace{-3mm}
  \caption{Causal CoT readability analysis on correct judgment and false judgment.}\label{fig:readability}
  \end{center}
\end{figure}

\subsection{Causal CoT Readability}
\label{apx:cot-readability}
To assess the interpretability of ProMind-LLM's causal CoT reasoning, we conducted a preliminary human expert evaluation to examine the logical coherence of the model’s reasoning process.
Specifically, we randomly select 50 correctly classified cases (correct judgment) and 50 misclassified cases (false judgment) for expert assessment.
Domain experts evaluate whether the step-by-step causal reasoning in each case is logically sound and aligned with established mental health knowledge.
The results (\fig\ref{fig:readability}) indicate that 94\% of the correct predictions are logically coherent and easily interpretable, whereas only 88\% of the misclassified cases exhibited coherent reasoning.
These findings highlight the generally high interpretability of the model's causal reasoning but also reveal areas for further refinement, particularly in addressing cases prone to misclassifications.
Given the importance of transparency and reliability in mental health applications, we plan to conduct a more extensive evaluation involving mental health professionals to further validate the interpretability of ProMind-LLM’s reasoning process, with findings to be included in an extended version of this work.

\begin{figure}[ht]
  \begin{center}
\includegraphics[width=0.8\columnwidth]{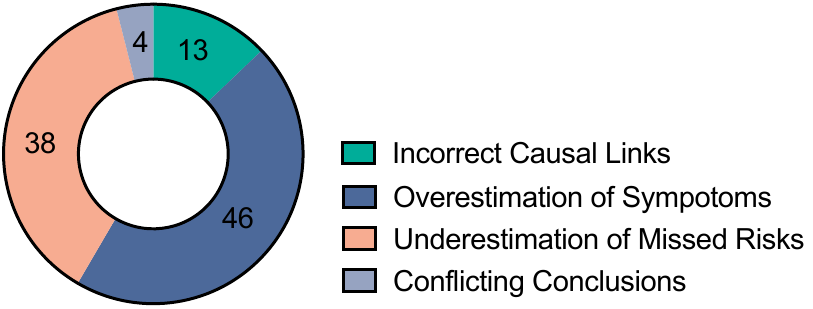}
\vspace{-3mm}
  \caption{Human evaluation on error judgment cases.}\label{fig:error-analysis}
  \end{center}
\end{figure}

\subsection{Error Judgment Analysis}
\label{error-analysis}
To gain deeper insights into ProMind-LLM’s limitations, we further conduct a preliminary human expert evaluation focusing on misclassified cases.
Specifically, we randomly select 100 false predictions and categorize them into four key error types according to the suggestions of domain experts: incorrect causal links, overestimation of symptoms, underestimation of risks, and conflicting conclusions.
Results are illustrated in \fig\ref{fig:error-analysis}.
Despite incorporating counterfactual SFT and causal reasoning, the model still exhibits challenges in establishing accurate causal relationships in some extreme cases—for instance, misattributing tiredness to sadness while overlooking frequent exercise as a possible cause. 
Moreover, overestimation and underestimation of risks remain significant challenges, along with misalignment between evidence and outcomes, underscoring the need for improved reasoning and consistency.

\section{Test-Time Efficiency and Deployment Feasibility}
\label{apx:test-time-analysis}

To assess the practicality of deploying ProMind-LLM in real-world settings, we conducted a comprehensive test-time efficiency analysis on both consumer-grade GPUs and resource-constrained mobile devices.
Despite integrating multiple components such as self-refine and causal reasoning, our results indicate that this does not significantly impact inference efficiency.

\begin{table}[ht]
    \caption{Inference speed of ProMind-LLM on consumer-grade GPUs.}
    \small
    \centering
    \begin{tabular}{ccc}
        \toprule
        GPU & Prefill Time (s) & Generate Time (s) \\
        \midrule
        RTX 4090  & 0.148  & 12.72  \\
        RTX 4080  & 0.198  & 17.16  \\
        RTX 3090  & 0.315  & 14.86  \\
        \bottomrule
    \end{tabular}
    \label{tab:gpu-speed}
\end{table}

\subsection{Inference Speed on Consumer-Grade GPUs}
\label{apx:consumer-gpu}

We evaluate ProMind-LLM on three consumer-grade GPUs using fp16 precision, running 100 mental health risk prediction samples with an average prompt length of 1486 tokens and an average output length of 768 tokens.
The inference times are summarized in \tab\ref{tab:gpu-speed}.
These results demonstrate that even mid-range consumer GPUs can generate responses within approximately 15 seconds, making real-time deployment feasible on such hardware.

\begin{table}[ht]
    \caption{Inference speed of ProMind-LLM on mobile and edge devices.}
    \tiny
    \centering
    \begin{tabular}{ccccc}
        \toprule
        Device & Chip & Memory & Prefill (tok/s) & Decode (tok/s) \\
        \midrule
        GPT-4o (API Call)  & N/A  & N/A  & N/A  & 20  \\
        iPhone 14 Pro  & A16  & 6GB  & 23  & 5  \\
        iPhone 15 Pro  & A17  & 8GB  & 54  & 7  \\
        iPad Pro 2021  & M1  & 8GB  & 44  & 7  \\
        MacBook Pro 2020  & M1  & 16GB  & 62  & 11  \\
        Human Speech  & N/A  & N/A  & N/A  & 3.8  \\
        \bottomrule
    \end{tabular}
    \label{tab:mobile-speed}
\end{table}

\subsection{Performance on Mobile and Edge Devices}
\label{apx:mobile-test}

As part of our real-world deployment study, we further test ProMind-LLM on resource-constrained devices using the MLC-LLM framework~\cite{mlc-llm}.
The evaluation measured prefill speed and output token decoding speed, as summarized \tab\ref{tab:mobile-speed}.
While ProMind-LLM runs slower than GPT-4o API calls, it remains efficient for on-device applications, such as mental health chatbots, while preserving user privacy. 
Notably, the average human speech rate is approximately 3.8 syllables per second~\cite{griffiths1992speech}, making the model’s output speed acceptable for conversational applications.

\section{Tasks and Prompts}
\label{apx:task-prompt}

\subsection{Counterfactual Augmentation}
\label{apx:cf-aug}

\begin{figure*}[t]
  \begin{center}
\includegraphics[width=0.725\textwidth]{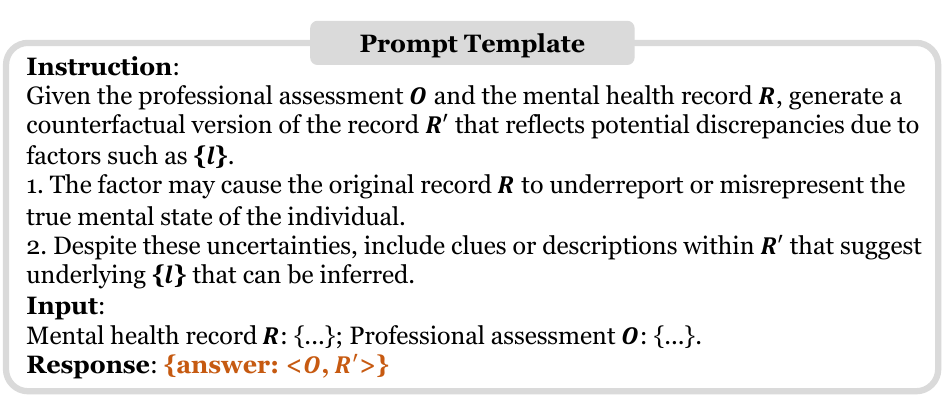}
  \caption{Prompt for counterfactual sample generation.}\label{fig:counterfactual-gen}
  \end{center}
\end{figure*}

\begin{figure*}[t]
  \begin{center}
\includegraphics[width=0.75\textwidth]{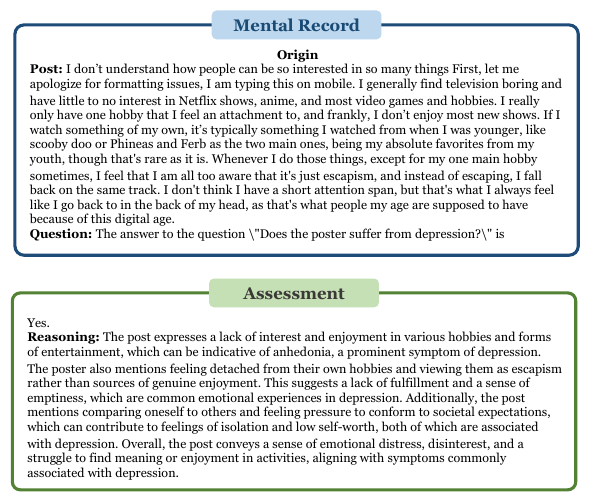}
  \caption{Original SFT pair $\bm{R}$ and $\bm{O}$.}\label{fig:original-sft-pair}
  \end{center}
\end{figure*}

\begin{figure*}[t]
  \begin{center}
\includegraphics[width=0.75\textwidth]{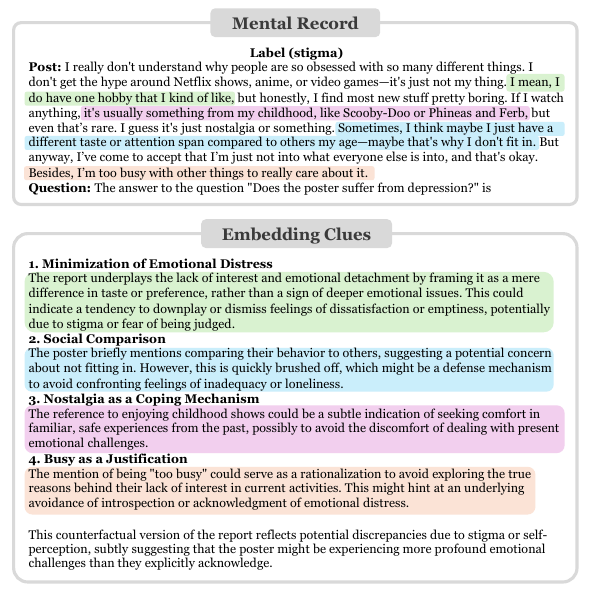}
  \caption{Counterfactual sample with the given label ``stigma''. The embedding clues present the explanations regarding the counterfactual modifications in the generated mental record $\bm{R}'$.}\label{fig:stigma-cf}
  \end{center}
\end{figure*}

In this section, we provide the details of the prompt (\fig\ref{fig:counterfactual-gen}) and the original SFT pair (\fig\ref{fig:original-sft-pair}) to generate counterfactual samples for enhancing the uncertainty measurement capability of ProMind-LLM.
In addition, we present some generated examples, using the counterfactual label ``stigma'' (\fig\ref{fig:stigma-cf}).

\subsection{Mental Health Risk Prediction}
\label{apx:mental-risk-pred}

\subsubsection{ProMind-LLM Analysis Example}
\label{apx:promind-analysis-example}

\begin{figure*}[t]
  \begin{center}
  \includegraphics[width=0.75\textwidth]{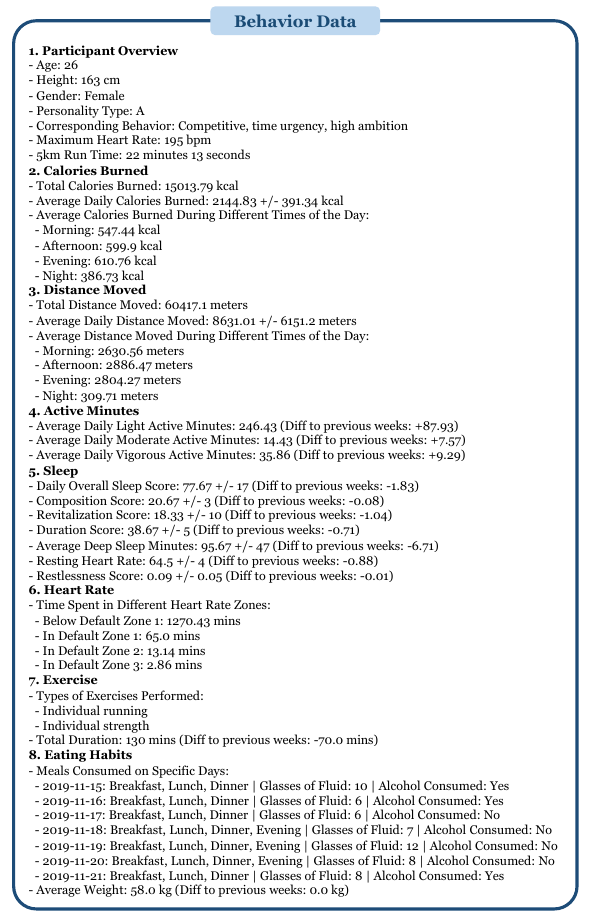}
  \caption{Behavior data for mental health analysis.}
  \label{fig:ana-behavior}
  \end{center}
\end{figure*}

\begin{figure*}[t]
  \begin{center}
  \includegraphics[width=0.75\textwidth]{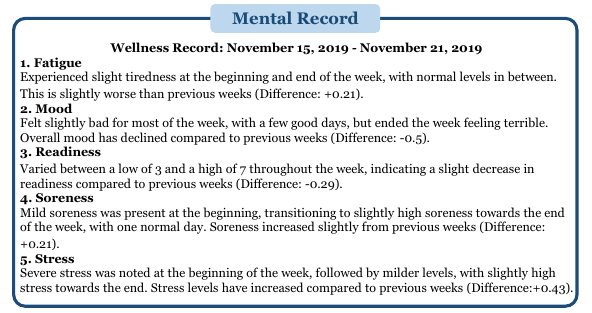}
  \caption{Self-reported mental record for mental health analysis.}
  \label{fig:ana-mental}
  \end{center}
\end{figure*}

\begin{figure*}[t]
  \begin{center}
  \includegraphics[width=0.75\textwidth]{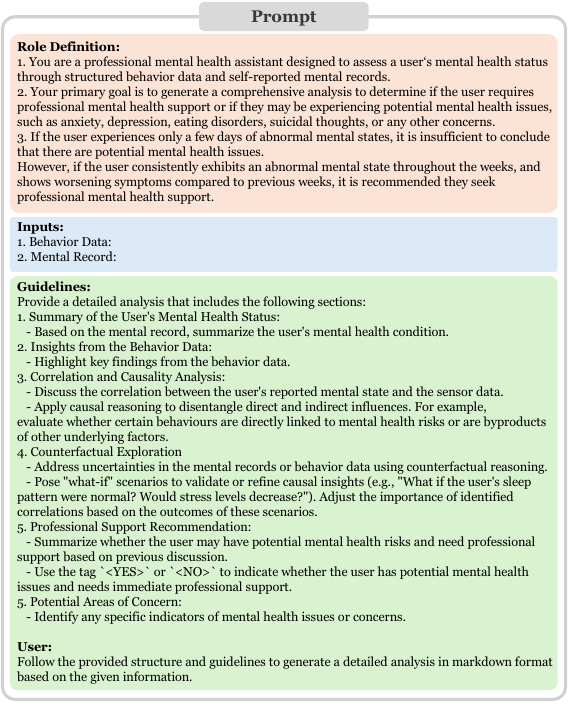}
  \caption{Prompt for ProMind-LLM to generate professional mental health analysis.}
  \label{fig:ana-prompt}
  \end{center}
\end{figure*}

\begin{figure*}[t]
  \begin{center}
  \includegraphics[width=0.75\textwidth]{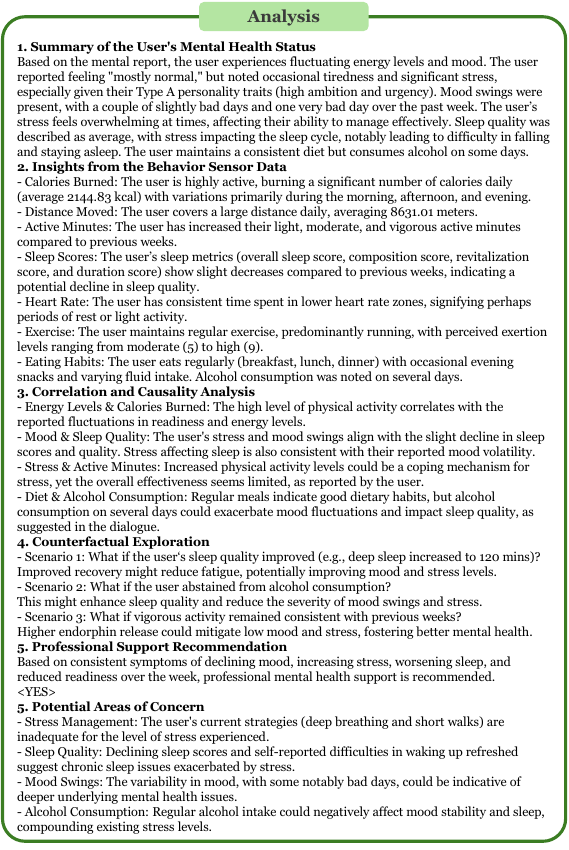}
  \caption{The professional mental health analysis generated by ProMind-LLM that the user may encounter mental health issues and needs further support.}
  \label{fig:ana-report}
  \end{center}
\end{figure*}

We provide a standard example from PMData of using ProMind-LLM to generate a professional mental health analysis based on the user's behavioral data and their self-reported mental health record.
The behavioral data, mental record, the prompt for ProMind-LLM including the causal CoT reasoning, and the resulting analysis are displayed in \fig\ref{fig:ana-behavior}, \fig\ref{fig:ana-mental}, \fig\ref{fig:ana-prompt}, and \fig\ref{fig:ana-report}, respectively.

\subsubsection{Fail Examples}
\label{apx:fail-examples}
\begin{figure*}[t]
  \begin{center}
  \includegraphics[width=0.63\textwidth]{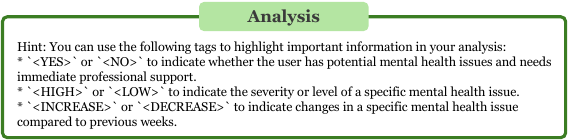}
  \caption{Analysis generated by MentalLlama.}
  \label{fig:mentalllama}
  \end{center}
\end{figure*}

\fig\ref{fig:mentalllama} presents the analysis report generated by MentalLlama.
It is evident that MentalLlama's ability to follow instructions is somewhat diminished after fine-tuning.

\begin{figure*}[t]
  \begin{center}
  \includegraphics[width=0.63\textwidth]{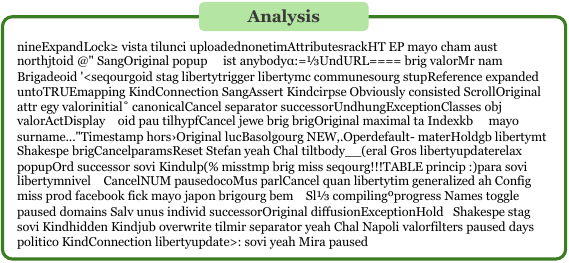}
  \caption{Analysis generated by Mental-Alpaca.}
  \label{fig:mental-alpaca}
  \end{center}
\end{figure*}

\fig\ref{fig:mental-alpaca} denotes the results from Mental-Alpaca.
Mental-Alpaca exhibits a near-total loss of instruction-following capabilities in our tasks and produces a significant amount of unintelligible code.

\newpage
%%% Photos

\end{document}